\documentclass{article}

\usepackage{arxiv}

\usepackage[utf8]{inputenc} 
\usepackage[T1]{fontenc}    
\usepackage{hyperref}       
\usepackage{url}            
\usepackage{booktabs}       
\usepackage{amsfonts}       
\usepackage{nicefrac}       
\usepackage{microtype}      
\usepackage{cleveref}       
\usepackage{lipsum}         
\usepackage{graphicx}
\usepackage{natbib}
\usepackage{doi, multirow, array}
\usepackage[symbol]{footmisc}

\title{An Extensible and Lightweight Unified Architecture for Demosaicing Pixel-bin Image Sensors}

\author{{Saurabh ~Kumar} \\
	AI Computational Imaging\\
	Samsung Research Institute Bangalore\\
	\texttt{saurabh.k1@samsung.com} \\
	\And
	{Nutan S.~Yenneti} \\
	AI Computational Imaging\\
	Samsung Research Institute Bangalore\\
	\texttt{ns.yenneti@samsung.com} \\
}

\renewcommand{\shorttitle}{An Extensible \&  Lightweight Unified Architecture for Demosaicing Pixel-bin Image Sensors}

\hypersetup{
pdftitle={An Extensible and Lightweight Unified Architecture for Demosaicing Pixel-bin Image Sensors},
pdfsubject={q-bio.NC, q-bio.QM},
pdfauthor={Saurabh ~Kumar, Nutan S.~Yenneti},
pdfkeywords={Computational Imaging, Camera ISP, Pixel-Bin Sensor, Color Filter Array, Demosaicing},
}

\begin{document}
\maketitle

\begin{abstract}
	Pixel-bin image sensors are becoming the default choice for smartphone cameras due to their resolution vs light-gathering trade-off. However, their larger inter-color separation compared to the Bayer color filter array (CFA) makes them challenging to demosaic. Furthermore, existing deep learning-based demosaicing methods are CFA-specific, requiring multiple individual models that take up precious onboard resources and demand larger development and maintenance efforts. In this work, we propose a modular unified architecture for demosaicing various pixel-bin sensors that provides higher image quality while being extensible and lightweight. Additionally, to enable plug-and-play operation, we introduce a learning-free CFA-identification module to detect the CFA type of raw data accurately.
\end{abstract}


\section{Introduction}
\label{sec:introduction}
Digital image sensors, being monochromatic discrete intensity measurement devices, employ a color filter array (CFA) to capture color photographs.
The CFA filters the incoming light onto the sensor into the three additive color primaries by sparsely sampling them spatially.
The process of interpolating the missing colors to estimate the full color at each pixel from these monochrome captures is called Demosaicing \cite{li2008image}. 
With the rapid rise in mobile photography, there is an increasing demand for higher image resolution and quality.
However, mobile cameras are heavily constrained by their onboard real estate and cost.
This has driven the already small mobile image sensors to accommodate higher pixel counts and, in turn, led to the shrinking of individual pixels themselves.
However, small pixels capture less light, leading to a lower image quality, especially in low-light scenarios.
The emergence of pixel-bin sensors addresses this challenge by offering a resolution vs light-gathering ability trade-off.

Pixel-Bin CFAs started with using a grid of $2 \times 2$ homogeneous color cells in place of a single Bayer CFA color cell and are referred to as the Quad or Tetra CFA \cite{barna2013method, chu2006improving}.
The technology moved forward towards Nona ($3 \times 3$) and quickly to Hexadeca ($4 \times 4$) CFA, also referred to as Tetra$^2$ or QxQ CFAs, which can be found in the latest Smartphones.
A visualization of color filter arrangements in various CFA types is shown in Figure \ref{fig:architecture}.
Based on the scene illumination, pixels corresponding to each homogeneous color cell grid can be binned.
For instance, even the small pixels can acquire sufficient light in a brightly lit scene, allowing for a high-resolution capture.
In low-light scenarios, homogeneous color cells are binned to gather more light and reduce noise while trading off the resolution.
Due to the increased inter-color gap, traditional methods \cite{li2008image} fail in demosaicing pixel-bin CFAs.

Lately, deep neural networks have been employed for demosaicing and offer improved image quality with faster reconstruction \cite{santos2025isp}.
Individual models have been proposed for Tetra \cite{a2021beyond, yang2022mipi, wu2023joint, zheng2024quad}, Nona \cite{kim2021recent, Sharif2021SAGANAS} and Hexadeca CFAs \cite{cho2023pynet}.
However, these approaches are CFA-specific by design.
In a real-world scenario, a pixel-bin sensor can output different CFA types, and a separate model is required to demosaic each of them.
Having multiple models onboard is challenging on resource-constrained devices like smartphones, as it requires storing and loading each of them independently, apart from the increased development and maintenance efforts.
Furthermore, a unified model allows for a common post-processing routine, enabling a consistent visual quality across CFA types.

Only two prior unified methods exist; KLAP by Lee et al. \cite{lee2023efficient} employs discriminative filters \cite{park2023all} and Tedla et al. \cite{tedla2025examining} use pixel shuffle with a residual network.
Both explicitly require the input CFA type, limiting their plug-and-play operation and propose large models not well-suited for onboard deployment.
Tedla et al. do not handle QxQ CFA and their extensibility is challenging.
We show a comparison with KLAP as only their implementation is available that handles all four CFA types.
In this paper, we address the above challenges, by proposing a CFA-ID module and a unified network architecture built to be lightweight and extensible while achieving a higher image quality.

Our key contributions in this work are: 
\textbf{1)} We propose an efficient and extensible unified neural architecture to demosaic various CFA patterns with its training methodology.
\textbf{2)} A lightweight learning-free CFA Identification module is proposed to enable plug-and-play operation.
\textbf{3)} We evaluate the proposed method on multiple synthetic datasets and present a new real sensor raw dataset and an evaluation on it as well. We also perform performance benchmarking of unified demosaicing models on state-of-the-art consumer hardware.

\section{Proposed Method}
\label{sec:proposedMethod}

\begin{figure}
    \includegraphics[width=\columnwidth]{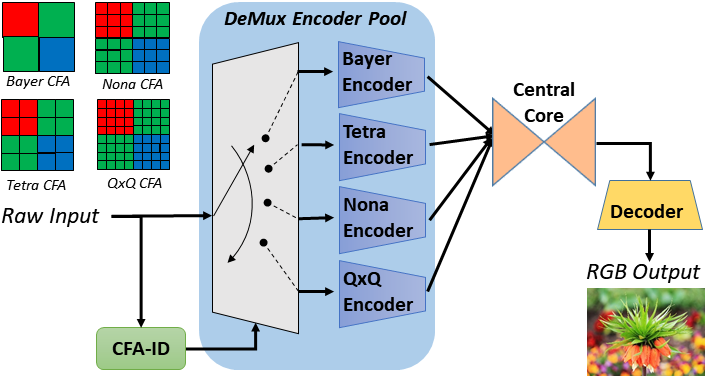}
    \caption{Overview of the proposed architecture and a visual of various pixel-bin CFAs. A given raw input is encoded using the Encoder selected by the CFA-ID module into a common latent space. These encodings are processed by the Central Core and finally decoded into an RGB image by the Decoder.}
    \label{fig:architecture}
\end{figure}

We propose a novel unified neural network architecture inspired by multi-view learning and digital system design that can demosaic various CFA patterns.
An overview of the proposed architecture is provided in Figure \ref{fig:architecture} that consists of four modules, namely a CFA Identification module, DeMux Encoder Pool, a Core, and a Decoder, which altogether take a CFA raw sensor image and output a demosaiced RGB image.

\subsection{Architecture Details}

\begin{figure}
    \begin{tabular}{m{5.4cm} m{5.4cm}}
        \multicolumn{2}{c}{\includegraphics[width=\columnwidth]{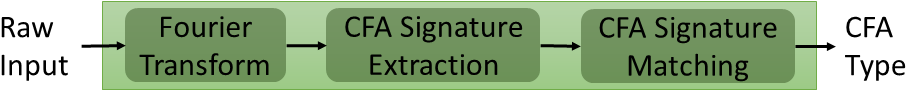}} \\
        \multicolumn{2}{c}{\footnotesize{(a)}} \\
        \includegraphics[width=0.63\columnwidth]{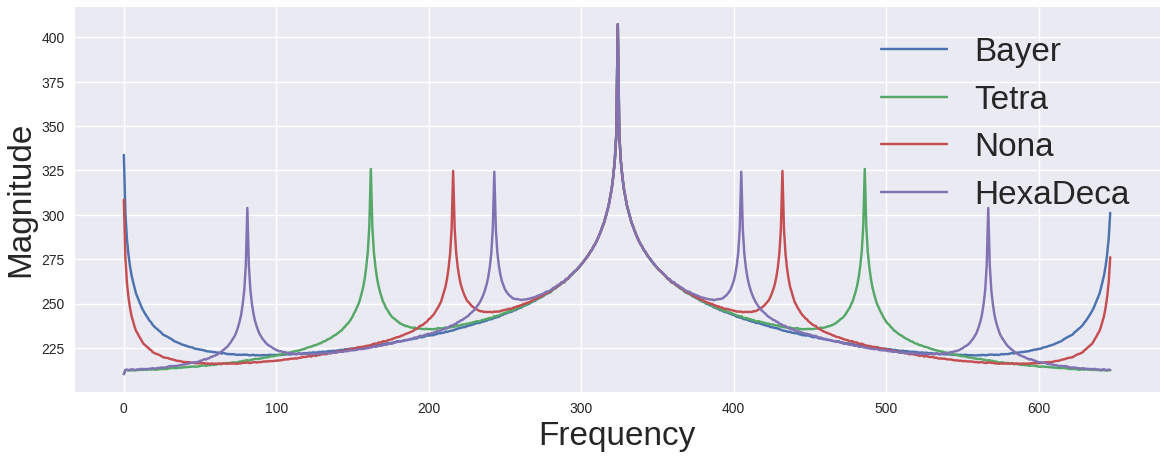} 
        & \includegraphics[width=0.33\columnwidth]{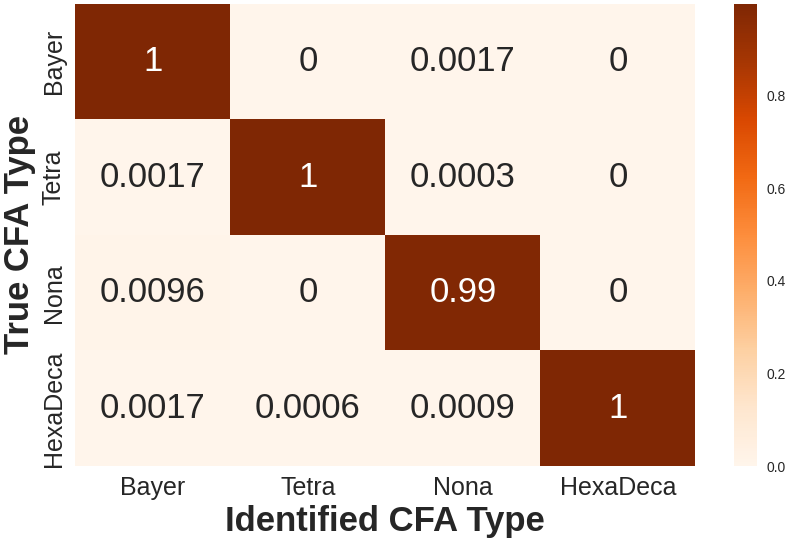} \\
        \centering \footnotesize{(b)} & 
        \centering \footnotesize{(c)} \\
    \end{tabular}
    \caption{Proposed CFA-Identification module : (a) Constituent blocks, (b) Representative CFA signatures, (c) Confusion matrix of the module's CFA identification performance.}
    \label{fig:CFAID}
\end{figure}

\textit{1) CFA Identification Module:}
We propose a novel CFA-ID module to identify the CFA type of the incoming sensor raw images in an efficient and learning-free way that enables the proposed unified model to work in a plug-and-play manner.
It consists of 3 parts; Fourier Transform, Signature Extractor, and Signature Matcher (Fig. \ref{fig:CFAID}(a)).
Given a raw image $I_{raw} \in \mathbb{R}^{M \times M}$, we first compute its 2D Fourier transform $\mathcal{F}$.
The next module computes the mean of the middle row and column of $|\mathcal{F}|$ to obtain the CFA Signature $\mathcal{S}$ as follows 
\begin{equation}
    \mathcal{S} = (|\mathcal{F}|_{[M/2,:]} + |\mathcal{F}|_{[:,M/2]})/2
\end{equation}
Finally, the Signature Matcher computes the correlation coefficients $\rho_{i}$ of the extracted CFA Signature of the input with each entry in the pre-computed CFA Signature Bank $\mathbf{\Sigma}$.
The largest $\rho_{i}$ indicates the predicted CFA type $c^*$ that is passed to the subsequent stages via select lines and computed as follows
\begin{equation}
    c^* = \arg\max_i (corr(\mathcal{S}, \mathbf{\Sigma_i)}) \textnormal{ } \forall \textnormal{ } i \in [B, T, N, H]
    \label{eq:maxCorr}
\end{equation}
where $\mathbf{\Sigma_i}$ denotes the CFA Signature Bank entries that are the representative signatures of each of the CFA types in consideration, which in our case are Bayer, Tetra, Nona, and Hexadeca, denoted by B, T, N, and H, respectively.
These representative CFA signatures are obtained by processing the synthetic raw training dataset using the Fourier Transform and Signature Extraction sub-modules and averaging the results for respective CFAs.
Figure \ref{fig:CFAID}(b) shows the computed representative CFA signatures for each CFA type.
This module can be easily extended to incorporate new CFAs by pre-computing their CFA Signature and including them in the CFA Signature Bank for matching.

\textit{2) DeMux Encoder Pool:}
The second stage is the DeMux Encoder Pool, which comprises a De-multiplexer and an Encoder for each CFA pattern in consideration, which in this work are: Bayer, Tetra, Nona, and Hexadeca.
Each Encoders comprise eight convolutional layers of kernel size $3 \times 3$ followed by Rectified Linear Units (ReLU) activations.
The encoders take a single channel input of the corresponding CFA types and encode them to a common latent space $\mathcal{L}$ of 64-channel features.
Depending on the inputs to the select lines from the CFA-ID module, an appropriate encoder encodes the input to its corresponding latent features.
This module is also extendable to new CFAs by adding their corresponding encoders and adapting the de-multiplexer.

\begin{table*}
    \centering
    \caption{Quantitative comparison of individual and unified demosaicing methods with proposed approach on synthetic datasets.}
    \resizebox{\textwidth}{!}{%
    \begin{tabular}{cc|ccc|ccc|ccc|ccc}
        \hline
         & & \multicolumn{3}{|c}{DIV2k} & \multicolumn{3}{|c}{BSD100} & \multicolumn{3}{|c}{URBAN100} & \multicolumn{3}{|c}{KODAK} \\
        CFA &   Method  &   PSNR$\uparrow$    &   SSIM$\uparrow$    &   LPIPS$\downarrow$   &   PSNR$\uparrow$    &   SSIM$\uparrow$    &   LPIPS$\downarrow$   &   PSNR$\uparrow$    &   SSIM$\uparrow$    &   LPIPS$\downarrow$ &   PSNR$\uparrow$    &   SSIM$\uparrow$    &   LPIPS$\downarrow$  \\
        \hline
        \multirow{3}*{Bayer}
            & PIPNet \cite{a2021beyond}     & 40.44   &   0.9903   &   0.0099  
                                            & 42.29   &   0.9949   &   0.0081 
                                            & 38.62   &   0.9901   &   0.0079
                                            & 40.62   &   0.9913   &   0.0079 \\
            & KLAP \cite{lee2023efficient}  & 41.26   &   0.9749   &   0.0556 
                                            & 39.75   &   0.9745   &   0.0396 
                                            & 38.04   &   0.9761   &   0.0275
                                            & 38.98   &   0.9718   &   0.0528 \\
            & Proposed  & \textbf{45.33}    &   \textbf{0.9950}   &   \textbf{0.0060} 
                    & \textbf{42.63}   &   \textbf{0.9937}   &   \textbf{0.0048} 
                    & \textbf{39.84}   &   \textbf{0.9936}   &   \textbf{0.0040}
                    & \textbf{40.87}   &   \textbf{0.9939}   &   \textbf{0.0059} \\
        \hline
        \multirow{3}*{Tetra}
            & PIPNet \cite{a2021beyond}   & 37.36   &   0.9837   &   0.0214 
                                            & 39.18   &   0.9894   &   0.0169   
                                            & 35.86   &   0.9845   &   0.0163
                                            & 37.35   &   0.9857   &   0.0172 \\
            & KLAP \cite{lee2023efficient}      & 40.71   &   0.9736   &   0.0572 
                                            & 39.23   &   0.9731   &   0.0418  
                                            & 37.64   &   0.9752   &   0.0284
                                            & 38.56   &   0.9706   &   0.0539 \\
            & Proposed  & \textbf{43.88}   &   \textbf{0.9923}   &   \textbf{0.0465} 
                    & \textbf{42.63}   &   \textbf{0.9937}   &   \textbf{0.0048}
                    & \textbf{39.13}   &   \textbf{0.9912}   &   \textbf{0.0049}
                    & \textbf{40.03}   &   \textbf{0.9879}   &   \textbf{0.0068} \\
        \hline
        \multirow{3}*{Nona}
            & SAGAN \cite{Sharif2021SAGANAS}   & 40.07   &   0.9712   &   0.0611 
                        & 38.72   &   0.9717   &   0.0587  
                        & 37.04   &   0.9746   &   0.0323 
                        & 38.21   &   0.9703   &   0.649 \\
            & KLAP \cite{lee2023efficient}     & 40.38   &   0.9734   &   0.0565 
                                            & 38.86   &   0.9728   &   0.0424 
                                            & 37.21   &   0.9751   &   0.0285
                                            & 38.34   &   0.9709   &   0.0550 \\
            & Proposed  & \textbf{42.74}   &   \textbf{0.9902}   &   \textbf{0.0214} 
                    & \textbf{40.20}   &   \textbf{0.9873}   &   \textbf{0.0150}
                    & \textbf{38.34}   &   \textbf{0.9890}   &   \textbf{0.0119}
                    & \textbf{39.19}   &   \textbf{0.9848}   &   \textbf{0.0194} \\
        \hline
        \multirow{3}*{Q$\times$Q}
            & PyNetQ$\times$Q \cite{cho2023pynet}  & 31.67   &   0.9483   &   0.0541
                                                    & 30.06   &   0.9335   &   0.0646
                                                    & 30.24   &   0.9408   &   0.0624
                                                    & 30.24   &   0.9408   &   0.0624 \\
            & KLAP \cite{lee2023efficient}              & 40.34   &   0.9728   &   0.0582 
                                            & 38.99   &   0.9728   &   0.0427 
                                            & 37.35   &   0.9747   &   0.0287
                                            & 38.37   &   0.9702   &   0.0548 \\
            & Proposed  & \textbf{43.32}    &   \textbf{0.9920} &   \textbf{0.0104} 
                    & \textbf{41.00}   &   \textbf{0.9885}   &   \textbf{0.0095}
                    & \textbf{38.74}   &   \textbf{0.9904}   &   \textbf{0.0072}
                    & \textbf{39.68}   &   \textbf{0.9855}   &   \textbf{0.0127} \\
        \hline
       \end{tabular}
       }
    \label{tab:quantitativeSynth}
\end{table*}

\textit{3) Central Core:}
The third stage is the Central Core, which performs unified processing of the encoded features of various CFA inputs in a common latent space $\mathcal{L}$.
It is of a U-Net architecture and comprises its Encoder, Decoder sub-modules, and skip connections.
The encoder sub-modules of the Central Core consist of four convolution layers of kernel size $3 \times 3$ followed by ReLU activations.
The decoder sub-modules comprise five transposed-convolution layers of kernel size $3 \times 3$, also followed by ReLU activations.
The final output of the Central Core is a 64-channel latent representation similar to its input via a ReLU non-linearity.
The proposed architecture allows the Central Core to learn from inputs of various CFA types during training and can be seamlessly extended to more CFA types as needed.

\begin{figure}[t]
    \begin{tabular}{c | m{3cm} m{3cm} m{3cm} m{3cm}}
    \toprule 
      & \multicolumn{1}{c}{Bayer} & \multicolumn{1}{c}{Tetra} & \multicolumn{1}{c}{Nona} & \multicolumn{1}{c}{Q$\times$Q} \\
    \midrule
    Raw  & \includegraphics[width=0.20\columnwidth]{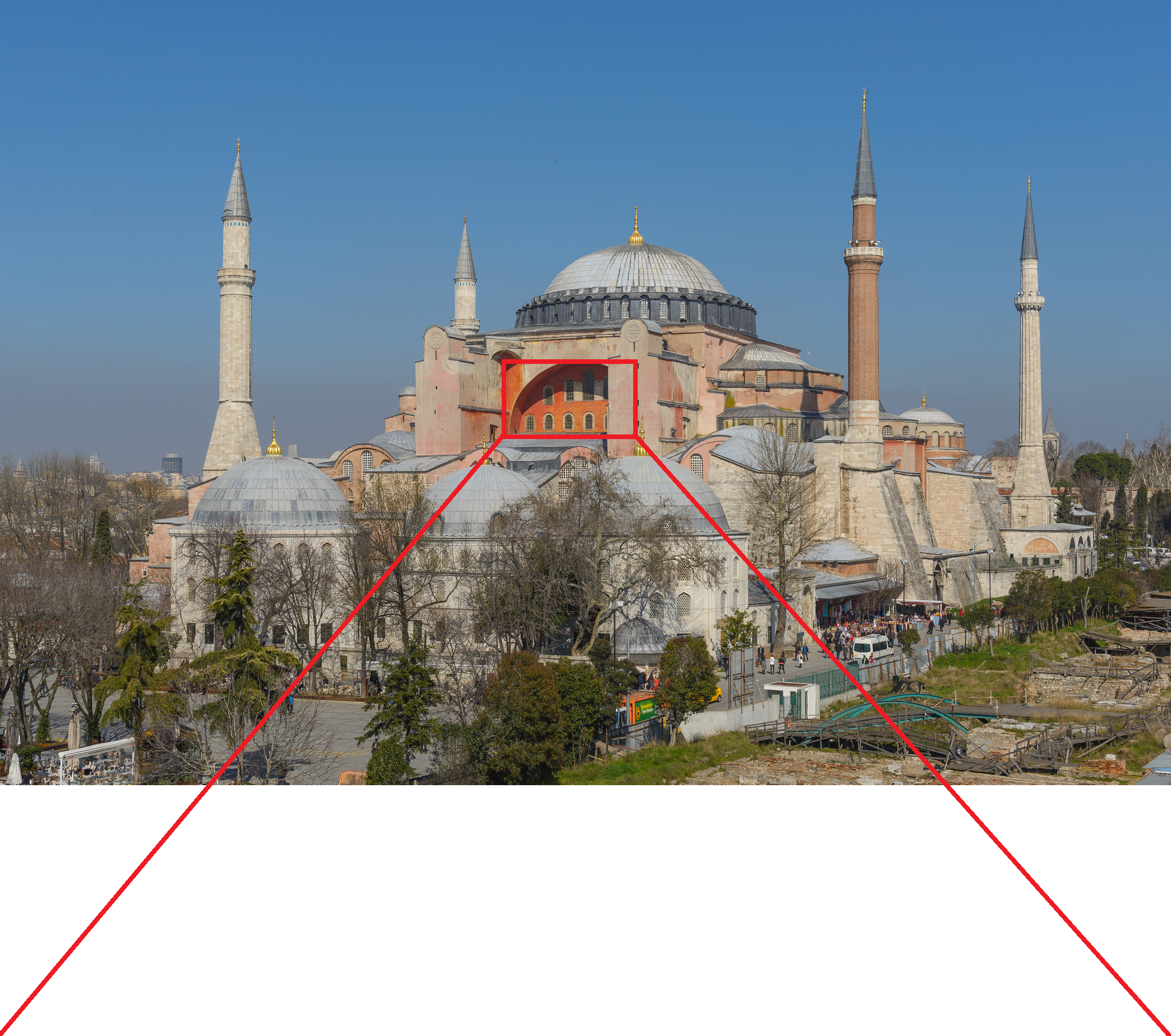}
        & \includegraphics[width=0.20\columnwidth]{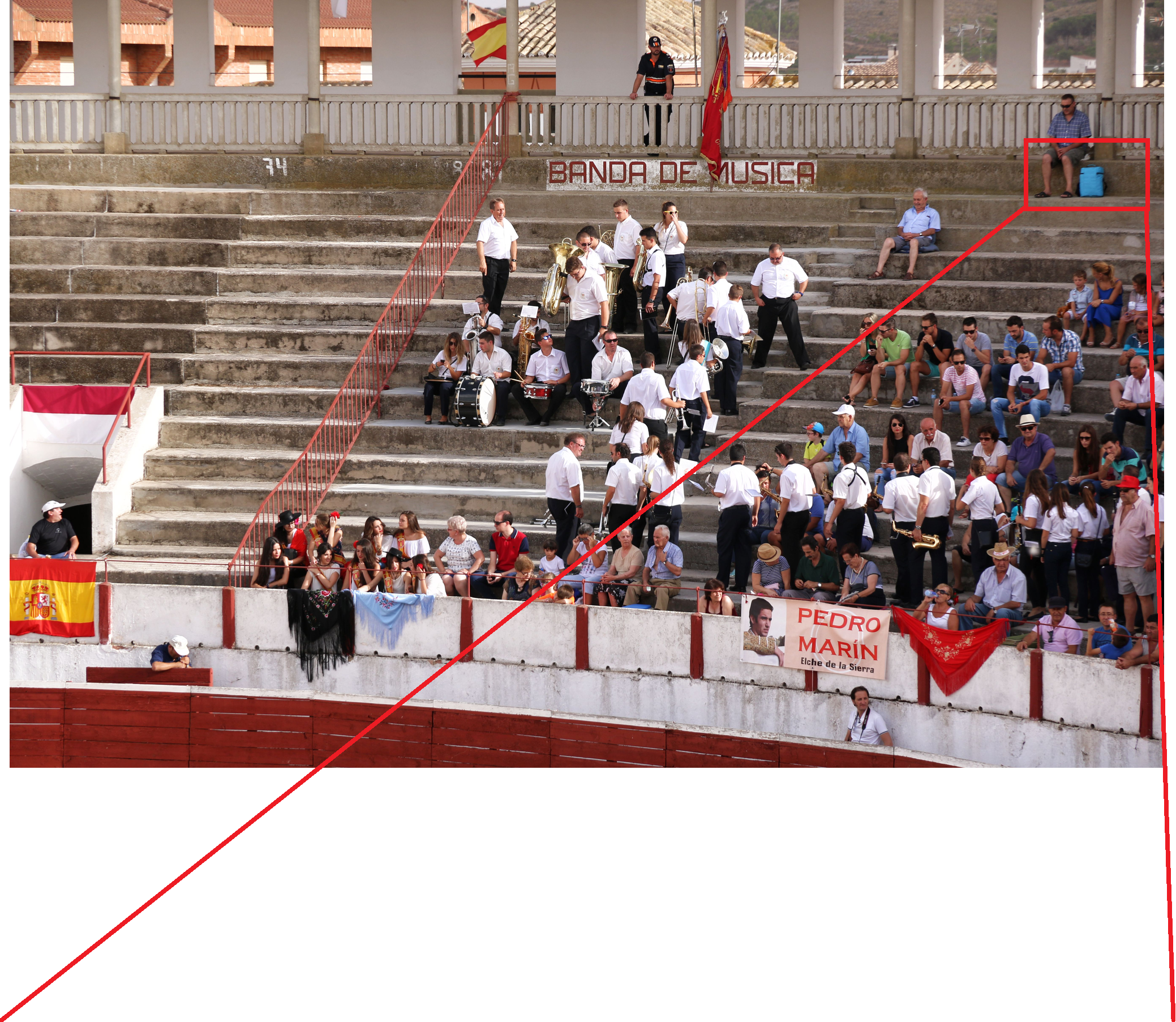}
        & \includegraphics[width=0.20\columnwidth]{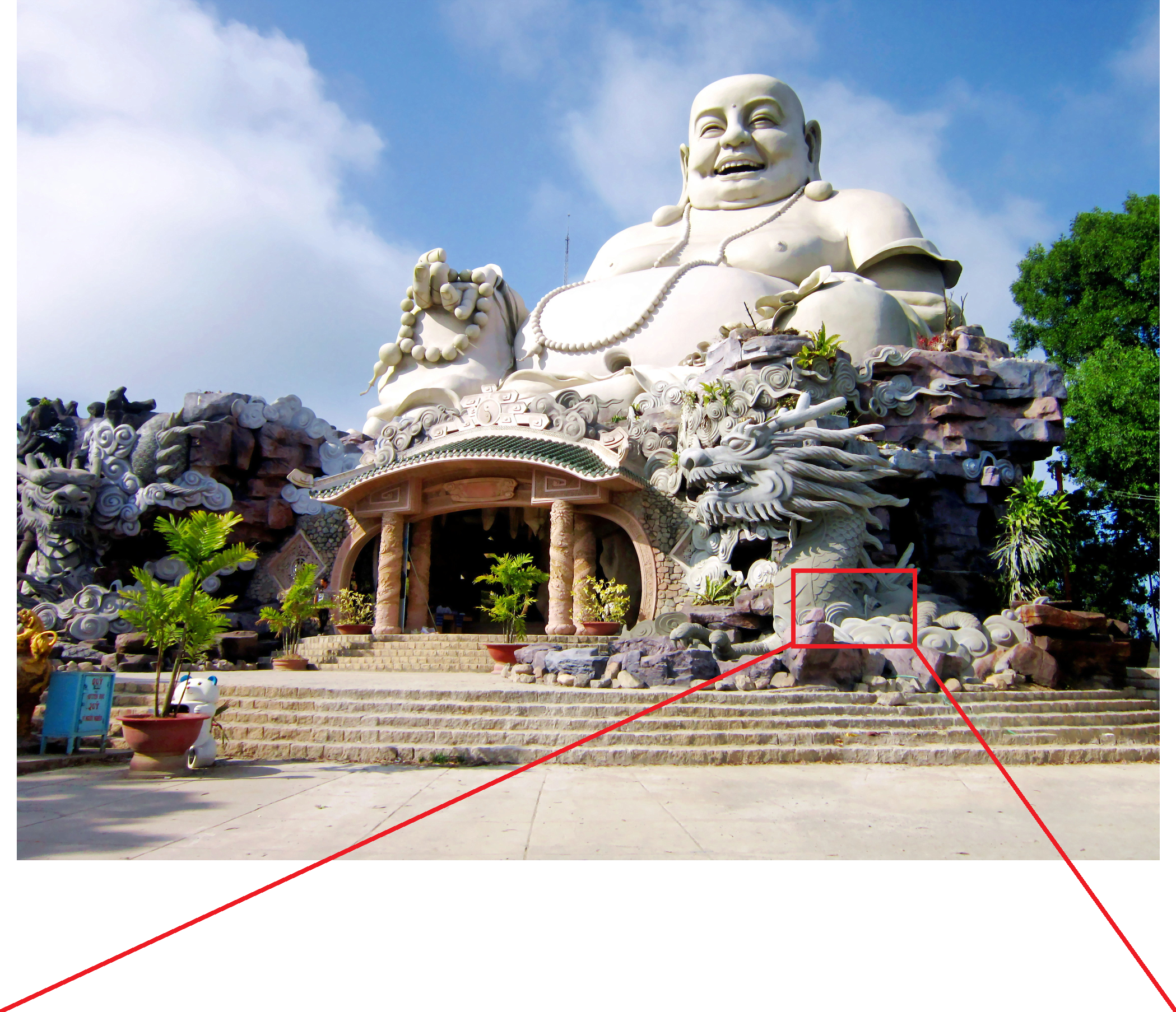} 
        & \includegraphics[width=0.20\columnwidth]{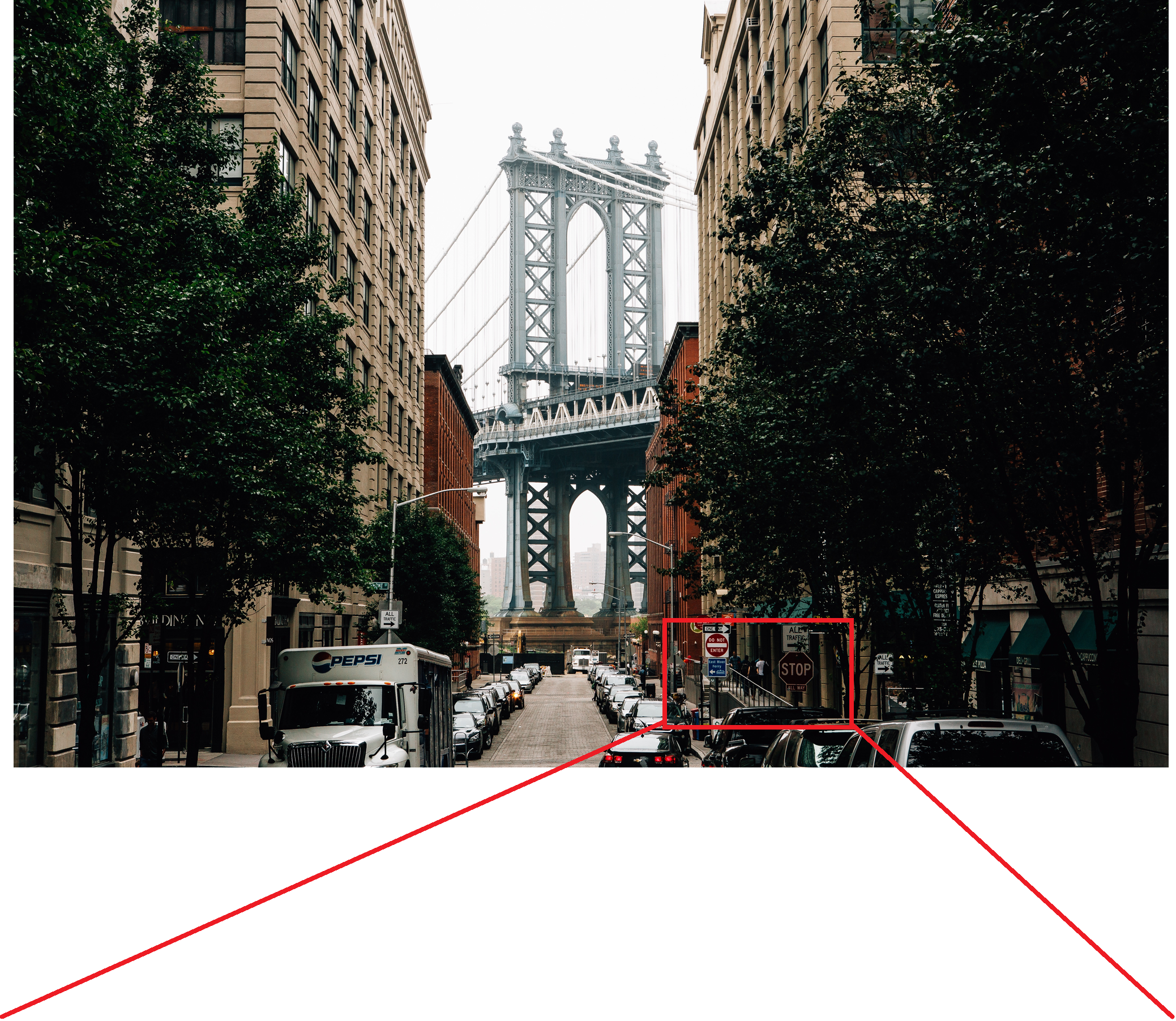} \\ 
    GT  & \includegraphics[width=0.20\columnwidth]{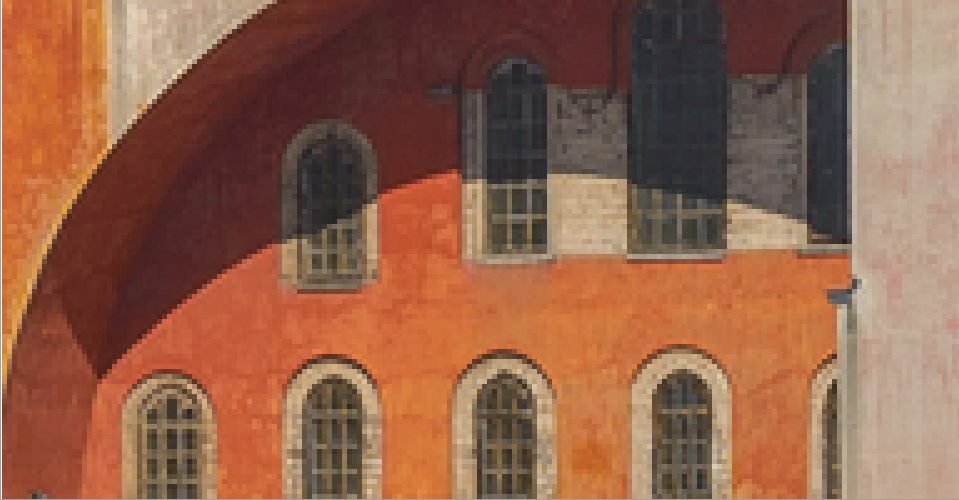}
        & \includegraphics[width=0.20\columnwidth]{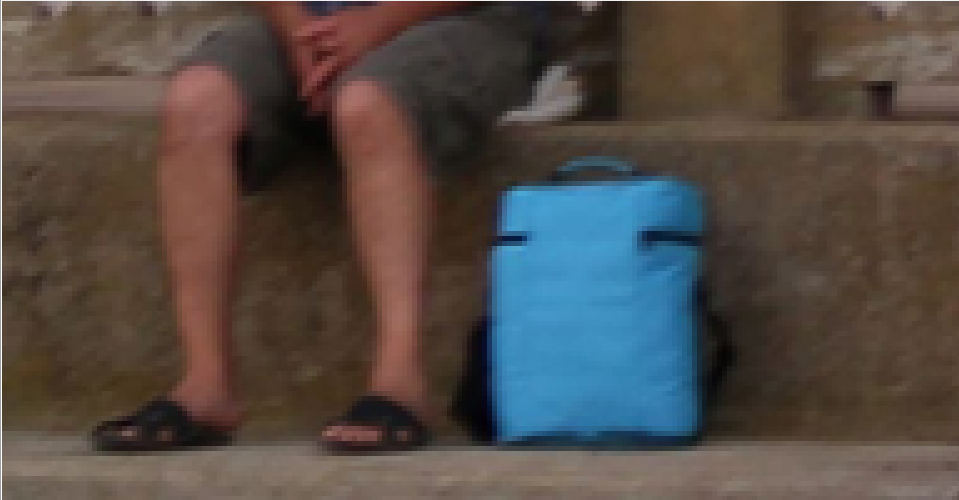}
        & \includegraphics[width=0.20\columnwidth]{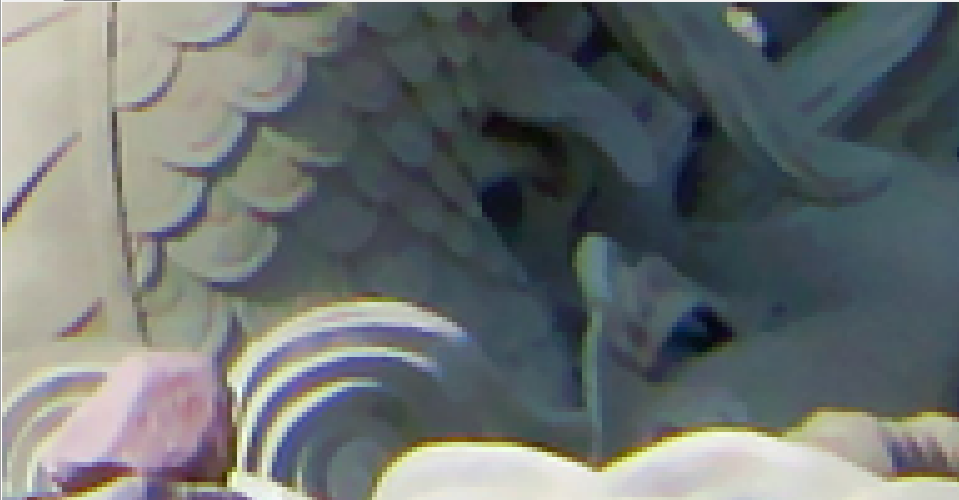} 
        & \includegraphics[width=0.20\columnwidth]{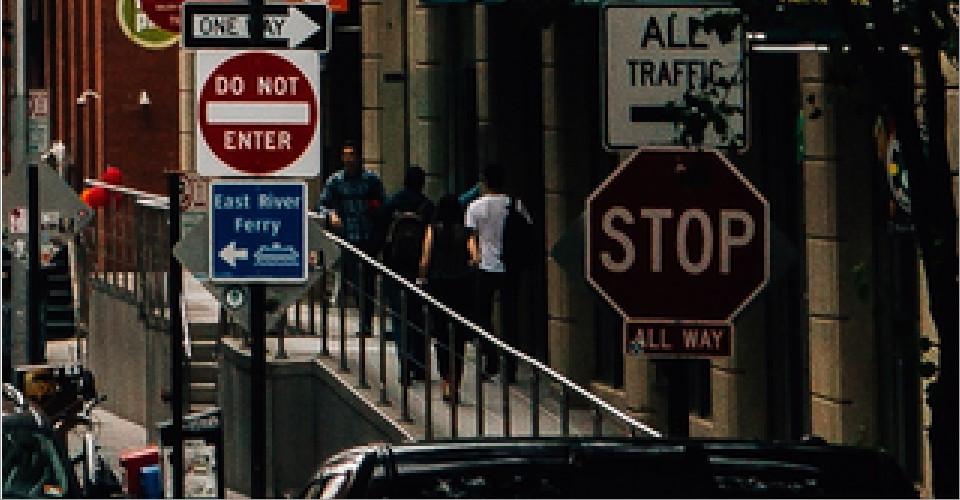} \\ 
    IM  & \includegraphics[width=0.20\columnwidth]{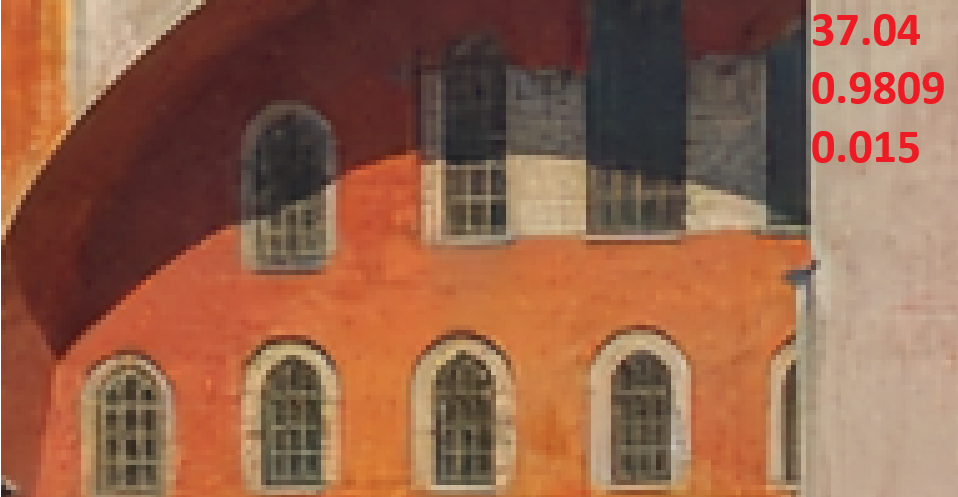} 
        & \includegraphics[width=0.20\columnwidth]{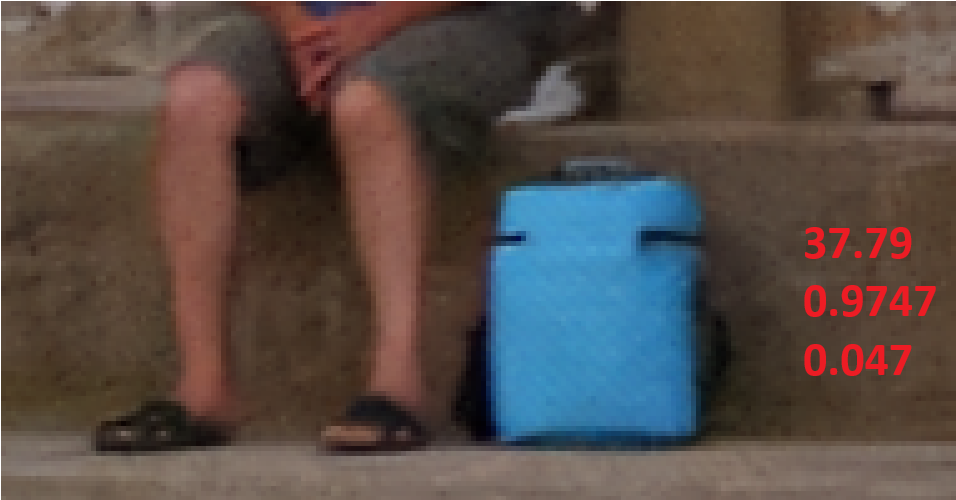}
        & \includegraphics[width=0.20\columnwidth]{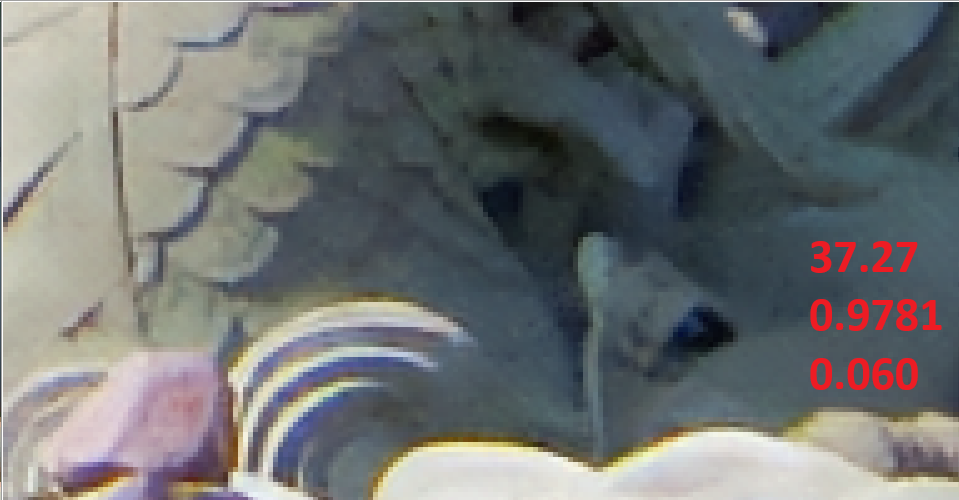}
        & \includegraphics[width=0.20\columnwidth]{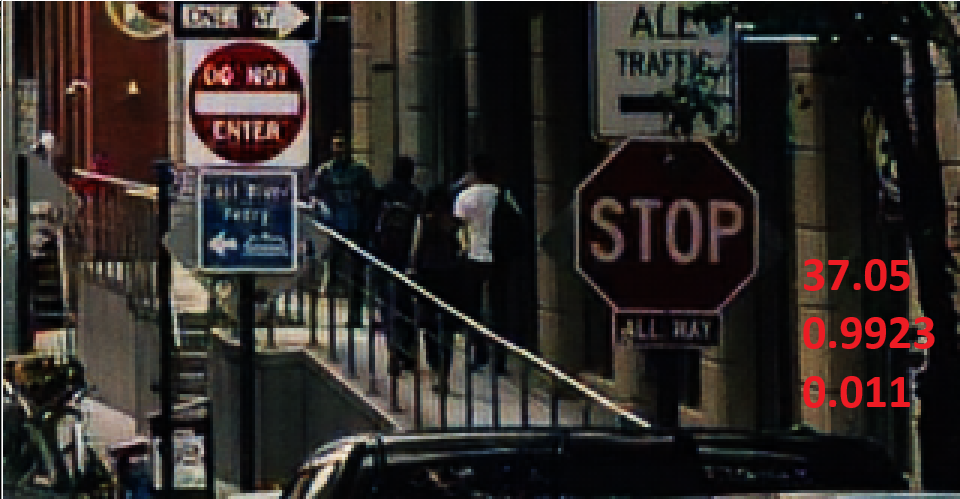} \\
    KLAP& \includegraphics[width=0.20\columnwidth]{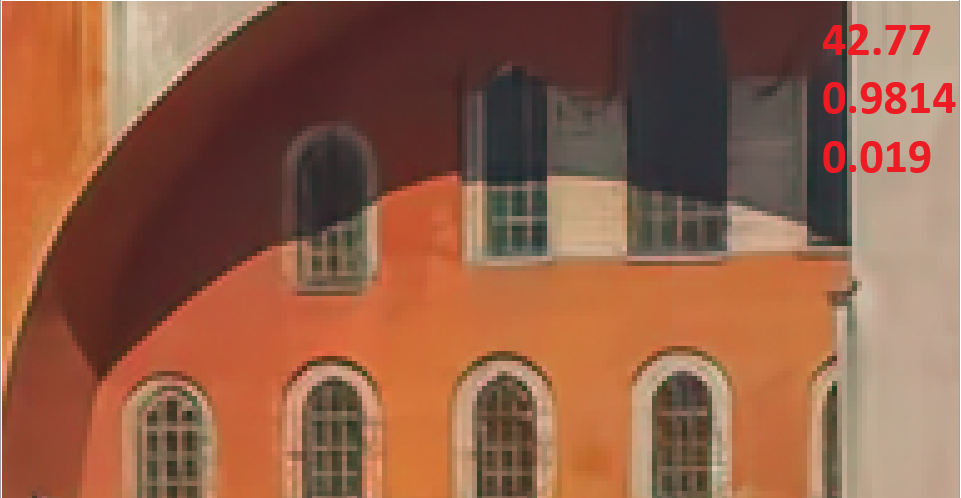} 
        & \includegraphics[width=0.20\columnwidth]{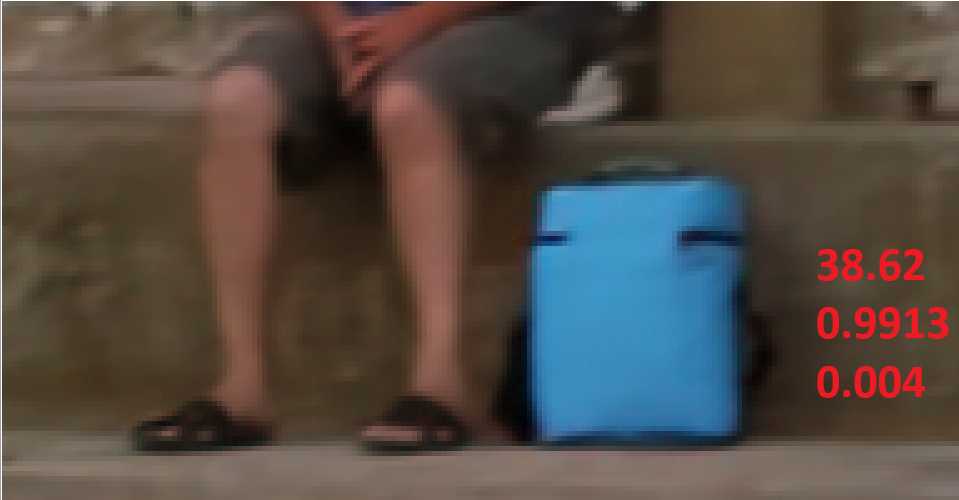} 
        & \includegraphics[width=0.20\columnwidth]{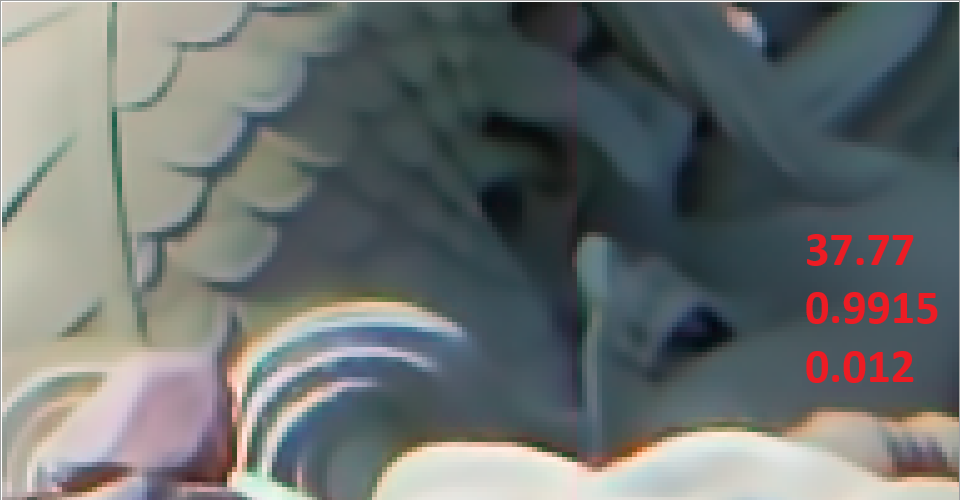}
        & \includegraphics[width=0.20\columnwidth]{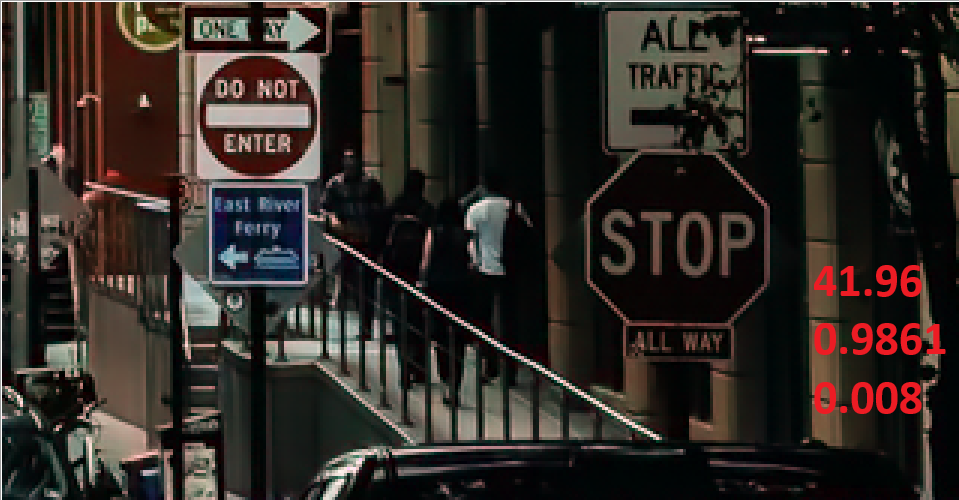} \\
    Proposed & \includegraphics[width=0.20\columnwidth]{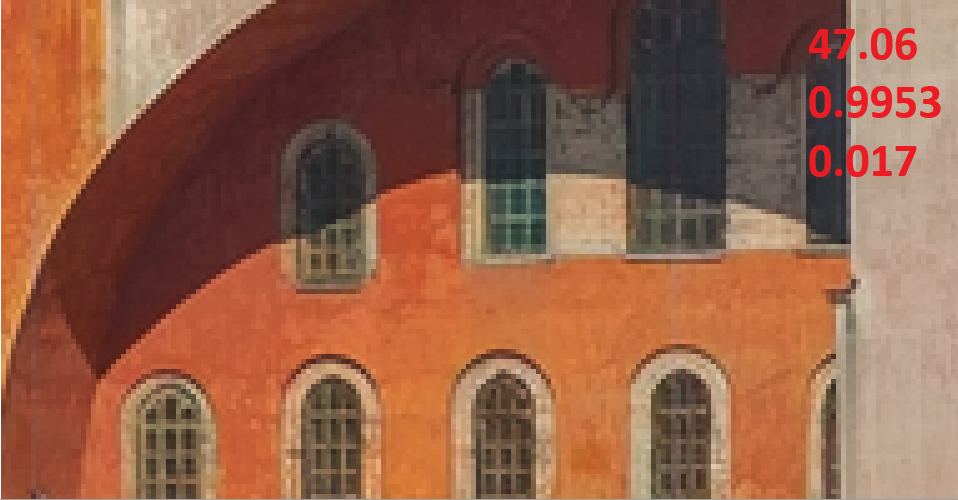}
        & \includegraphics[width=0.20\columnwidth]{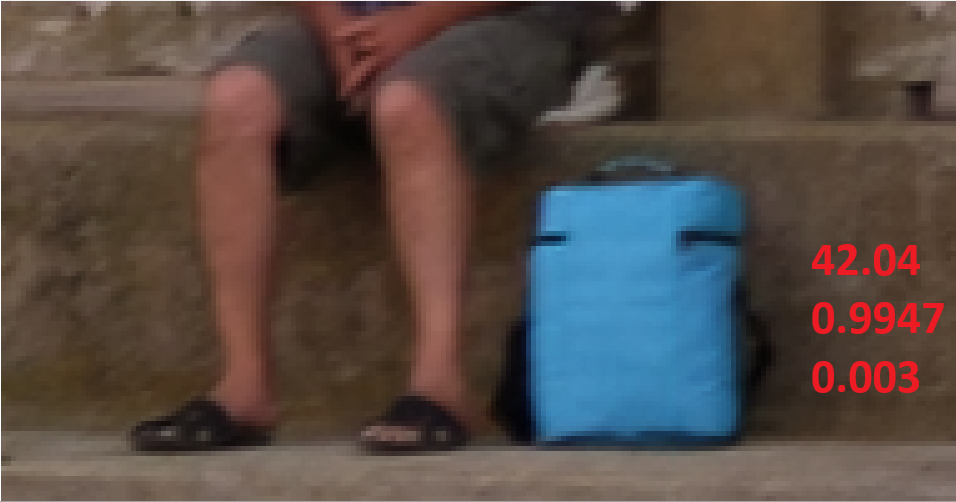} 
        & \includegraphics[width=0.20\columnwidth]{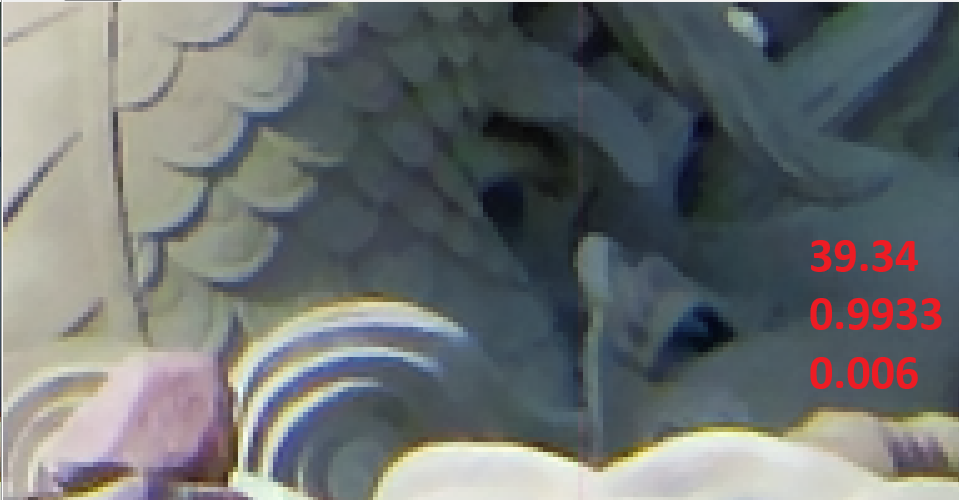}
        & \includegraphics[width=0.20\columnwidth]{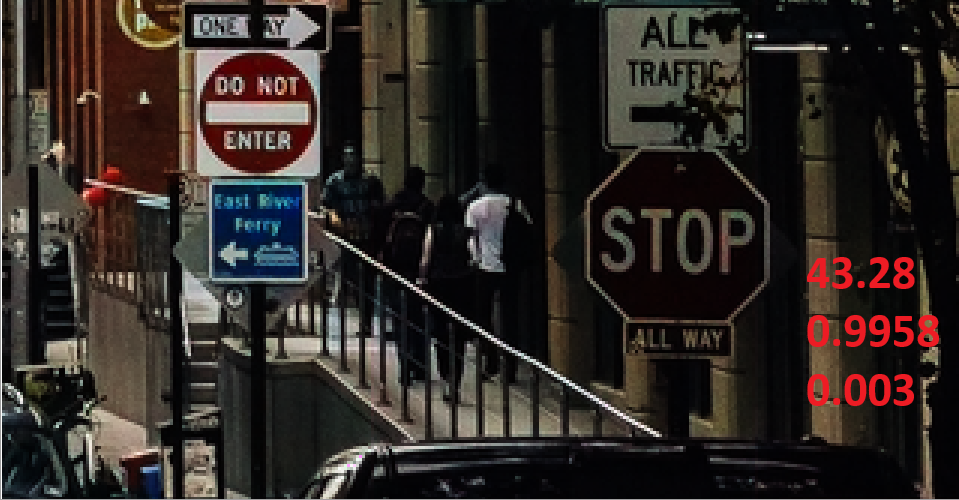} \\ 
    \bottomrule
    \end{tabular}
    \caption{Qualitative comparison of proposed approach with individual models (IM) and unified model (KLAP) on synthetic raw data. Each of the zoomed insets are annotated with the respective PSNR, SSIM and LPIPS metrics from top to bottom.}
    \label{fig:qualitativeSynth}
\end{figure}

\textit{4) Decoder:}
The final stage of our architecture is the Decoder module that takes in the processed latent features from the Central Core to reconstruct an RGB image.
The Decoder is composed of four convolution layers of $3 \times 3$ kernel size, first three are followed by ReLU activations and the fourth is by a Sigmoid non-linearity.
It takes a 64-channel latent representation in common latent space $\mathcal{L}$ as input, and the final convolution layer outputs a three channel representation that is our final demosaiced output RGB image.

\subsection{Training Methodology}
We perform a four-stage training of our unified model since four CFA patterns are considered for this paper.
In each stage, an input is passed through the CFA-ID module, and the corresponding Encoder is trained jointly with the Central Core and Decoder in an end-to-end supervised manner using the input and its respective ground truth RGB image.
An Encoder is selected randomly in each stage of training, while other Encoders remain frozen. 
We train the proposed model using Adam optimizer \cite{kingma2014adam} with a learning rate of $10^{-4}$.
We minimize a convex combination of L1 loss and Multi-scale structural similarity loss \cite{zhao2016loss} as our training objective and employ Xavier weight initialization \cite{kumar2017weight}.
The proposed architecture and training methodology facilitate seamless incorporation of new CFA types, making them extensible.

\section{Results}
\label{sec:results}

\begin{figure}
    \begin{tabular}{c | m{7cm} m{7cm}}
    \toprule 
        & \multicolumn{1}{c}{KLAP} & \multicolumn{1}{c}{Proposed}\\ 
    \midrule
    Bayer & \includegraphics[width=0.44\columnwidth]{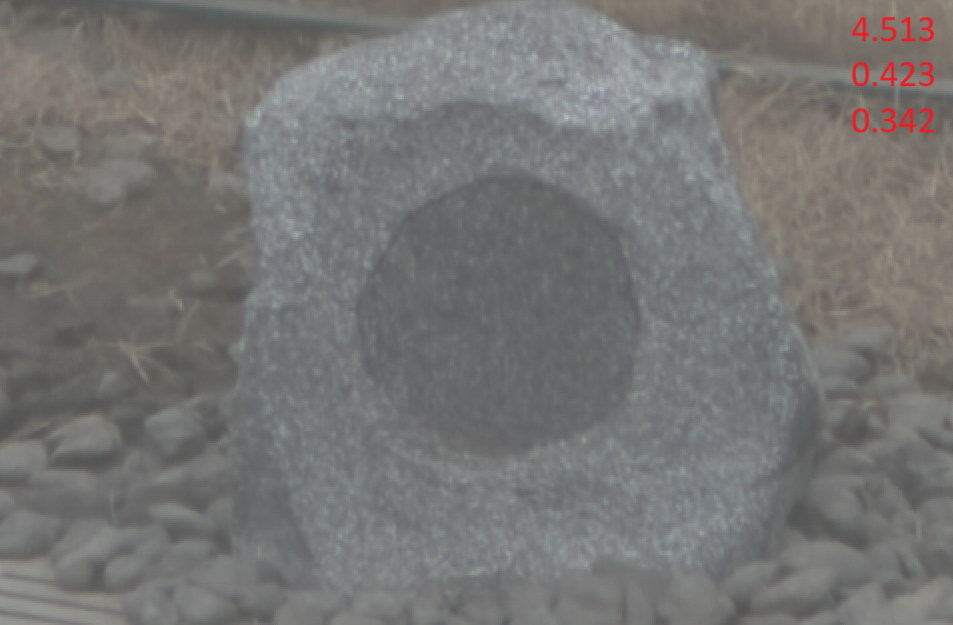}
        & \includegraphics[width=0.44\columnwidth]{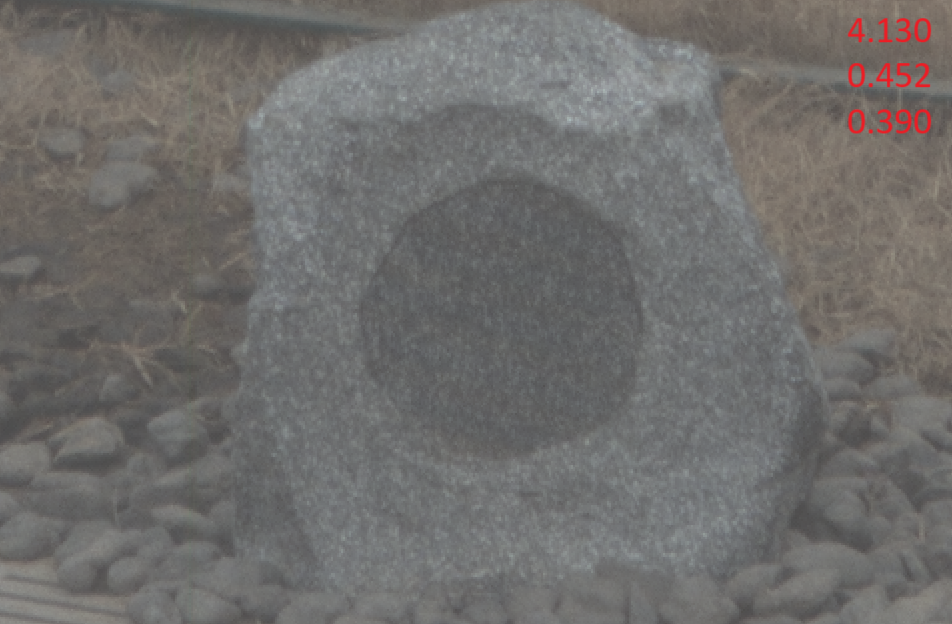} \\
    Tetra & \includegraphics[width=0.44\columnwidth]{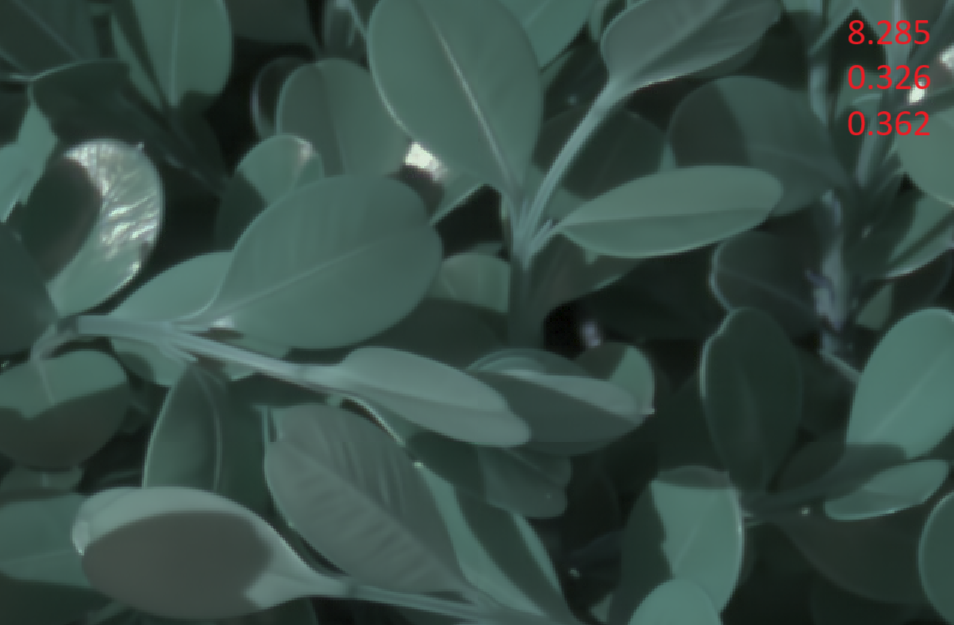}
        & \includegraphics[width=0.44\columnwidth]{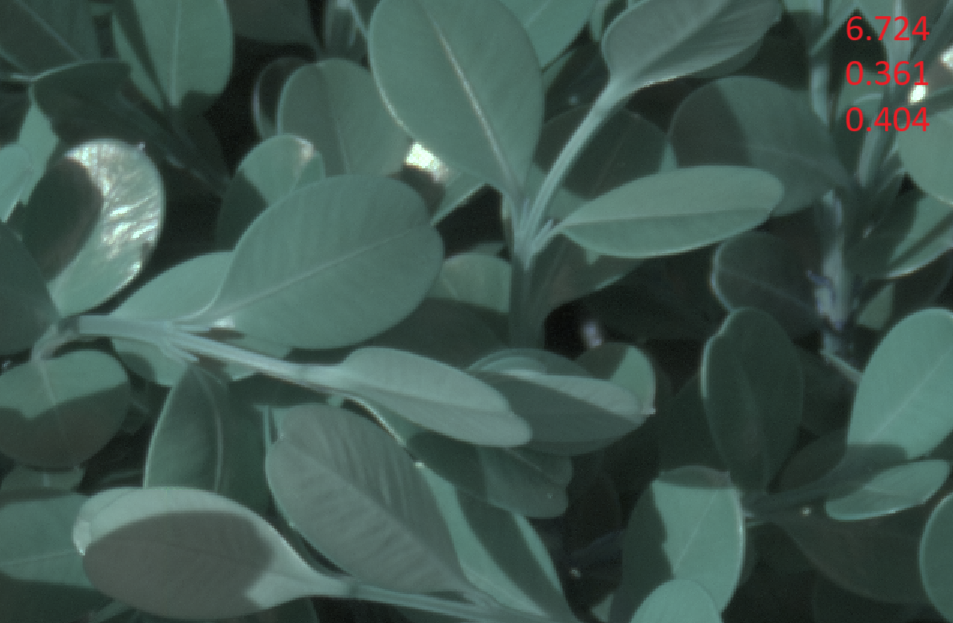} \\ 
    Nona & \includegraphics[width=0.44\columnwidth]{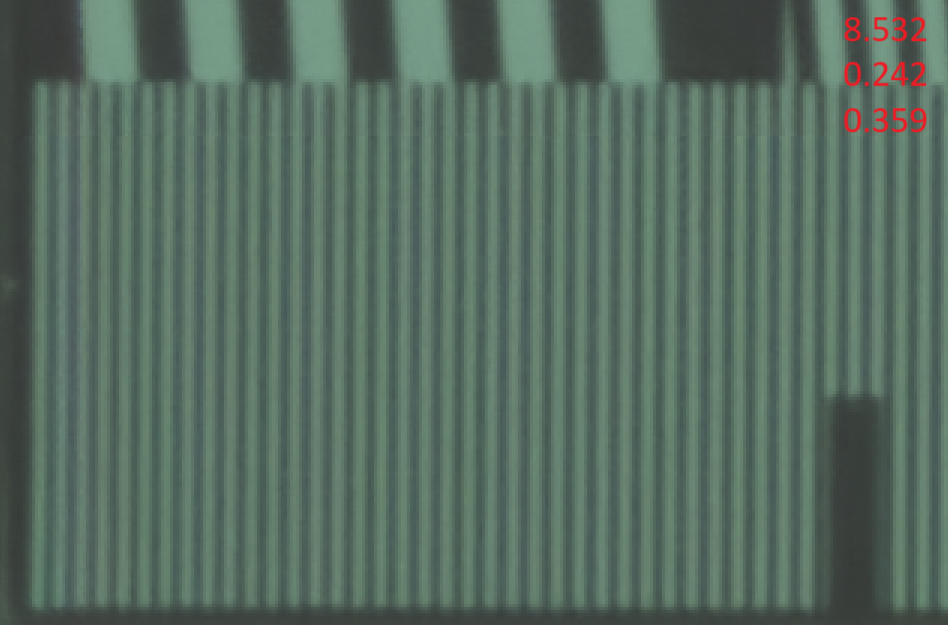} 
        & \includegraphics[width=0.44\columnwidth]{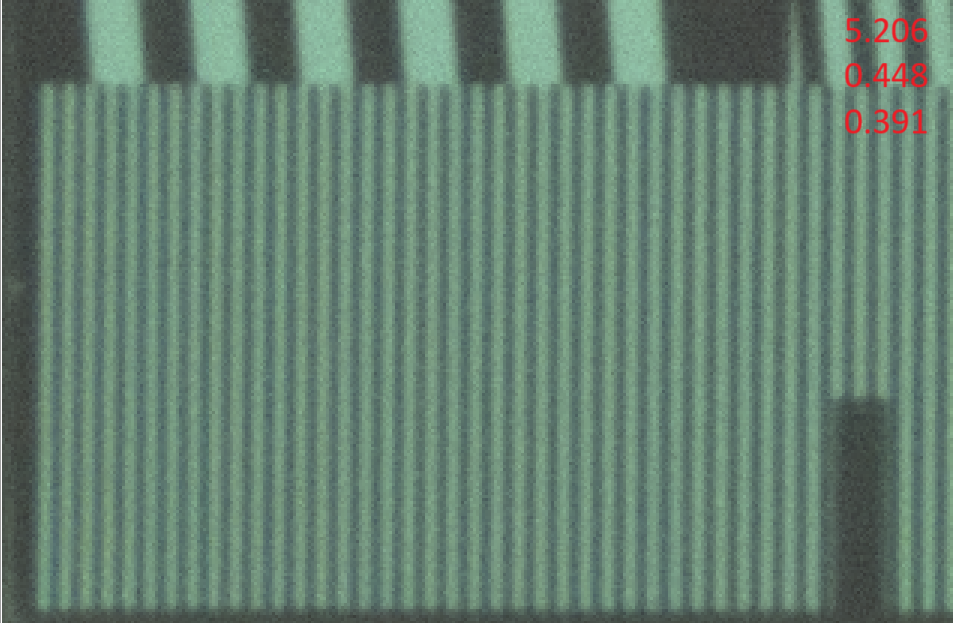} \\
    QxQ & \includegraphics[width=0.44\columnwidth]{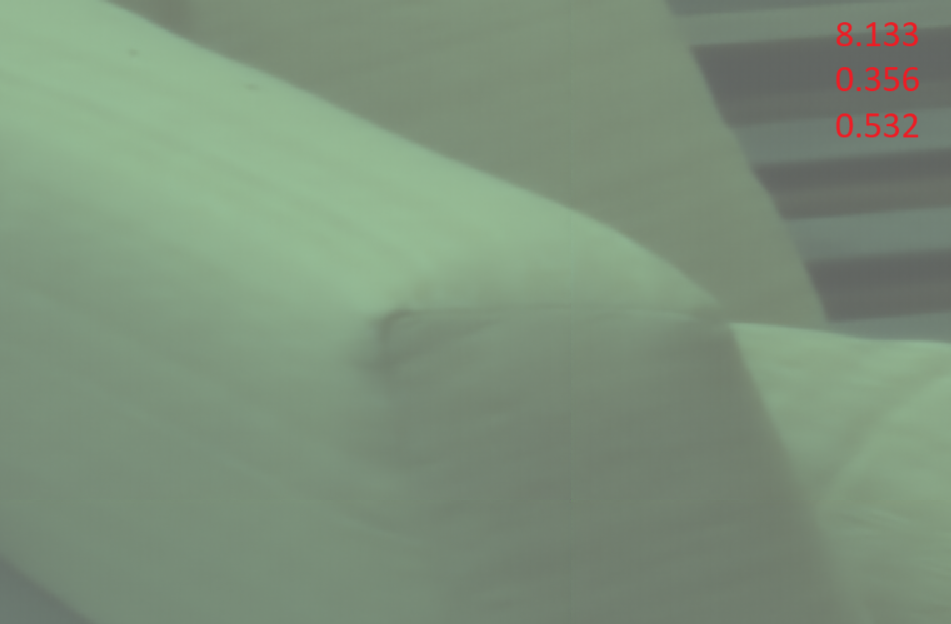}
        & \includegraphics[width=0.44\columnwidth]{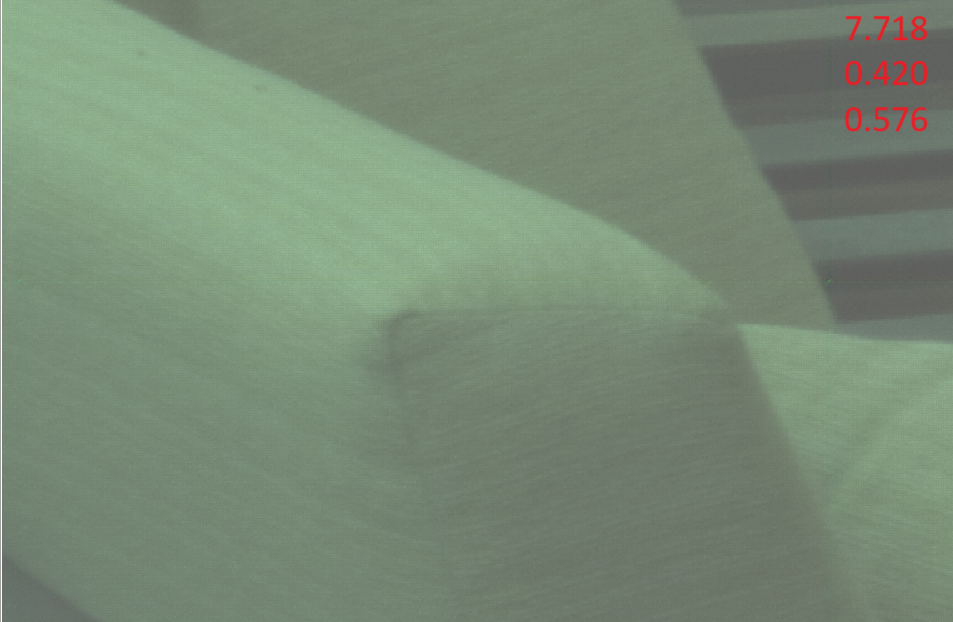} \\ 
    \bottomrule
    \end{tabular}
    \caption{Visual results of our method and KLAP on real data, with inset annotated with NIQE, ARNIQA \& CNNIQA metrics.}
    \label{fig:qualitativeReal}
\end{figure}

To demonstrate the effectiveness of our proposed method, we present qualitative and quantitative comparisons with existing individual and unified demosaicing approaches on both synthetic and real raw data.
For a fair comparison with prior work \cite{lee2023efficient, brooks2019unprocessing}, we employ their synthetic data generation approach of simulating an inverse ISP pipeline to process RGB images and obtain synthetic raw samples.
We use the DF2K dataset for training, with 800 samples from the DIV2K training set and 2650 samples from the Flickr2K dataset.

The prior works \cite{lee2023efficient, tedla2025examining} use private datasets for evaluation, limiting their reproducibility.
In this paper, we address this gap by using 4 public datasets for our synthetic data evaluations, namely, the DIV2K val set \cite{Agustsson_2017_CVPR_Workshops}, BSD100 \cite{MartinFTM01}, Urban100 \cite{huang2015single}, and Kodak.
Similarly, for real data evaluation, we capture Bayer, Tetra, Nona, and QxQ samples from Samsung cameras, which shall be made available for reproducibility.
No such dataset is currently available in the literature.

\subsection{Testing on Synthetic Raw Data}
We present a quantitative comparison of the proposed approach with previous state of the art methods in Table \ref{tab:quantitativeSynth} and a qualitative comparison in Figure \ref{fig:qualitativeSynth}.
The quantitative results demonstrate that the proposed method outperforms both the SoTA unified method \cite{lee2023efficient} and the corresponding individual models \cite{a2021beyond, cho2023pynet} for all CFAs by a significant margin across datasets.
The qualitative results also demonstrate that the proposed method provides a higher fidelity reconstruction than prior works.
In particular, we can see in the Bayer CFA results where the individual model (PIPNet \cite{a2021beyond}) leads to artifacts and the unified model (KLAP \cite{lee2023efficient}) leads to detail loss as compared to the proposed method.
A similar trend is visible in the results for Tetra and Nona CFAs, and it becomes more evident in the case of QxQ demosaicing, where the stop sign is significantly distorted in IM (PyNetQxQ \cite{cho2023pynet}), and KLAP leads to loss of details as compared to the proposed method.

\subsection{Testing on Real Raw Data}
To demonstrate our proposed approach's effectiveness and robustness, we test it on real raw captures from various CFA sensors to evaluate real-world performance shown in Figure \ref{fig:qualitativeReal}.
The same model weights as synthetic data evaluation are used here without retraining or fine-tuning.
Due to a lack of ground truth RGBs, we present no-reference image quality metrics, namely, NIQE \cite{mittal2012making}, ARNIQA \cite{agnolucci2024arniqa}, and CNNIQA \cite{kang2014convolutional}, .
Again, from both the qualitative and quantitative comparison, we can see that the proposed method translates well to the real data and provides an improved quality and detail preservation than KLAP \cite{lee2023efficient}. 
Our method also avoids color aberrations as seen in the results of Nona CFA in Fig. \ref{fig:qualitativeReal}.

\begin{table}
    \centering
    \caption{On-device performance on Snapdragon 8750 SoC}
    \begin{tabular}{c|cc}
        \hline
         Metric & KLAP & Ours \\
         \hline
        Parameter Count & 17.8 M\footnotemark / 85.05 M\footnotemark &  14.78 Million \\
        NPU Runtime     & DLC Conversion Fails    & 40.87 ms \\
        \hline
       \end{tabular}
    \label{tab:benchmarks}
\end{table}

\footnotetext{Reported in KLAP paper \cite{lee2023efficient}}
\footnotetext{Calculated from open-source implementation of KLAP at : https://github.com/ignoww/KLAP}

\subsection{Efficacy of CFA Identification Module}
We also study the identification performance of the proposed CFA-ID module by predicting the CFA type of our synthetic raw dataset.
A confusion matrix of the module's performance is presented in Figure \ref{fig:CFAID}(c), where we can observe that the proposed module can identify the CFA type of the input with very high accuracy of over 99\%.
We want to highlight that, despite its good performance, this module has limitations when the scene is monochrome or a flat region with no texture, like a picture of a clear sky.
On visual examination, we observed that the incorrectly identified samples in the confusion matrix consist only of these samples.

\subsection{On-Device Performance}
Along with PC simulations, we perform on-device benchmarking of the proposed model on the latest consumer hardware for real-world evaluation.
Unified demosaicing models have not been tested in the literature for their onboard deployment feasibility and performance.
To address this gap, we deploy our model on Qualcomm Snapdragon 8750 chipset, the results for which can be found in Table \ref{tab:benchmarks} for an input of size $648\times648$.
The proposed model comprises $\sim6\times$ fewer parameters than the SoTA unified model while readily converting for on-device execution.

\section{Conclusions}
\label{sec:conclusion}
In this paper, we proposed a unified architecture for demosaicing various pixel-bin CFAs that works in a plug-and-play manner and achieves a higher reconstruction quality than both unified and individual models while being significantly light-weight.
Furthermore, our model uses only fundamental blocks like convolutional layers and ReLU, allowing for simpler training and a hardware-friendly design than prior works.
Finally, the proposed architecture is designed from the ground up to be seamlessly extendable to new CFA patterns.

\bibliographystyle{unsrtnat}
\bibliography{refs}

\end{document}